\documentclass[11pt]{article}

\usepackage[final]{acl}
\usepackage{times}
\usepackage{latexsym}
\usepackage[T1]{fontenc}
\usepackage[utf8]{inputenc}
\usepackage{microtype}
\usepackage{inconsolata}
\usepackage{hyperref}
\usepackage{url}
\usepackage{booktabs}
\usepackage{amsfonts}
\usepackage{amsmath}
\usepackage{amssymb}
\usepackage{xcolor}
\usepackage{graphicx}
\usepackage{multirow}
\usepackage{enumitem}
\usepackage{caption}
\usepackage{algorithm}
\usepackage{algpseudocode}
\usepackage{tikz}
\usetikzlibrary{arrows.meta,positioning,shapes.geometric,calc,fit,backgrounds}

\title{Debate-to-Skill: Capability-Bound Process Supervision\\for Industrial Query-to-Agent Annotation}

\author{
 \textbf{Shiyu Zhang\textsuperscript{1}},
 \textbf{Leisheng Cheng\textsuperscript{1}},
 \textbf{Huifu Li\textsuperscript{1}}\thanks{Corresponding author: \href{mailto:lihuifufu@163.com}{lihuifufu@163.com}}
\\
 \textsuperscript{1}Baidu, Inc.
\\
 \texttt{zhsy12345689@gmail.com}, \texttt{chengleisheng@baidu.com}, \texttt{lihuifufu@163.com}
}

\begin{document}

\maketitle

\begin{abstract}
Industrial query-to-agent matching fails when topical relevance is mistaken for executable capability, especially on long-tail and boundary-sensitive requests. We formulate annotation as \emph{capability-bound process supervision} and instantiate it with Debate-to-Skill, which uses reusable decision principles, structured deliberation, verifier-based verdict extraction, and disagreement-driven refinement. On an industrial Query2Agent benchmark, we compare Debate-to-Skill with direct-label supervision, reasoning-SFT, and structural ablations. The results test whether gains come from supervising the capability-critical decision process itself, especially on grey-zone cases where semantic relatedness and executable capability diverge.
\end{abstract}

\section{Introduction}

\begin{figure*}[t]
\centering
\includegraphics[width=\textwidth]{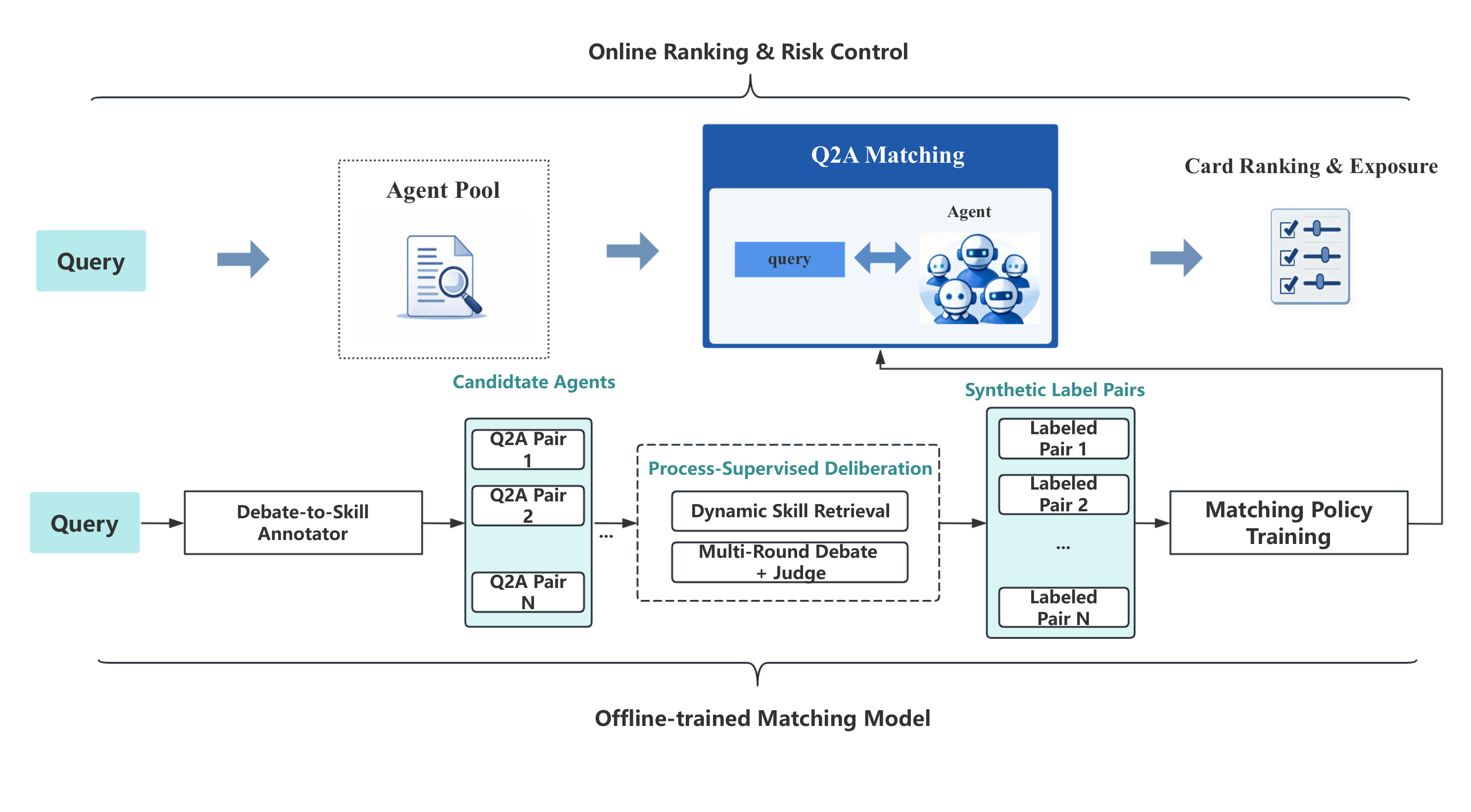}
\caption{Industrial Query2Agent workflow: online serving routes user requests to candidate agents, while the offline loop converts traffic and review signals into Debate-to-Skill supervision for routing-policy updates.}
\label{fig:baidu-workflow}
\end{figure*}

Large-scale search and agent platforms increasingly rely on query-to-agent matching to connect user requests with executable AI services. In industrial search and agent-serving surfaces, this decision directly affects card exposure, downstream engagement, and later policy learning. Recent advances in instruction following, tool use, and reasoning make agent serving increasingly practical \cite{wei2022flan,yao2023react,schick2023toolformer,asai2024selfrag}, but they do not remove the annotation bottleneck: production traffic is massive, long-tail, and boundary-sensitive, while high-quality expert labels remain scarce.

The main difficulty is that query-to-agent supervision is usually formulated as semantic relevance prediction, even though the costly production errors come from a different source. A candidate agent can look topically related yet still be invalid because it lacks the required tools, authority, real-time access, or service boundary to complete the task. Weak supervision and pseudo-labeling can amplify label volume \cite{burns2024weakstrong,shin2025weakstrong,bayesianweak2025}, and structured reasoning can make decisions more inspectable \cite{wei2022cot,wang2022selfconsistency,du2024,lightman2024letsverify}, but neither directly fixes this mismatch if the supervision target itself remains a final relevance label. We therefore argue that query-to-agent matching is not fundamentally a relevance problem. It is a capability verification problem under asymmetric risk.

We address this by formulating industrial annotation as \emph{capability-bound process supervision}. Instead of asking the model to emit a one-shot label, we supervise whether a query-agent pair passes a capability-critical decision process grounded in reusable principles, explicit support-vs.-failure evidence decomposition, structured verdict extraction, and disagreement-driven refinement. We instantiate this formulation with \textbf{Debate-to-Skill} and evaluate it against direct-label supervision, reasoning-SFT, and structural ablations on industrial Query2Agent benchmarks. The resulting experiments are designed to test a single claim: the main gains should come from supervising the capability-bound decision process itself, especially on grey-zone cases where semantic relatedness and executable capability diverge.

Our contributions are:
\begin{itemize}[leftmargin=1.5em]
\item We diagnose industrial query-to-agent annotation as a supervision-object problem: the dominant final-label formulation collapses semantic relatedness and executable capability, leading to systematic errors on grey-zone and boundary-sensitive cases.
\item We propose capability-bound process supervision as the correct target for this setting, and instantiate it with Debate-to-Skill, where reusable capability principles, evidence factorization, verifier extraction, and disagreement abstraction make the decision process observable and trainable.
\item We instantiate this process-supervision target through reasoning-SFT and verifier-based GRPO, showing that capability-bound traces provide a stronger learnable object than direct labels or generic reasoning.
\end{itemize}

\section{Problem Setting}

Each training example is a query-agent pair $x=(q,a)$, where $q$ is a user request from industrial traffic and $a$ is a candidate agent profile. The raw annotation space in our pipeline is ternary: $y_{\text{raw}}=2$ if the agent can satisfy the request, $y_{\text{raw}}=1$ if the pair is semantically related but capability-insufficient or risky, and $y_{\text{raw}}=0$ if it is not relevant. For training and standard evaluation, we collapse this into a binary target $y\in\{0,1\}$ by treating only $y_{\text{raw}}=2$ as positive. Unless otherwise stated, binary metrics are computed after this deployment collapse, while grey-zone diagnostics use the raw ternary verdict before collapse. This formulation mirrors the business objective: the central failure is not missing topical relatedness, but over-accepting grey-zone pairs whose apparent relevance does not survive capability-bound inspection. Our method is therefore designed to supervise the decision process that maps business-side capability risk into a learnable annotation target.

\paragraph{Grey-zone example.}
Consider the anonymized query ``My flight is delayed, help me rebook'' and a candidate ``Travel Assistant'' whose profile advertises travel consultation but exposes no ticketing, order-modification, inventory, or booking APIs. The pair is topically related, but it is not an executable match: rebooking requires access to the user's order and a transaction API. Debate-to-Skill retrieves principles about capability boundaries, real-time requirements, and consulting-versus-execution, supports semantic relevance, and then vetoes the pair on capability grounds. Gold labels are never included in evaluation or inference prompts.

\section{Method}

\subsection{Method Overview}

\begin{figure*}[t]
\centering
\includegraphics[width=\textwidth]{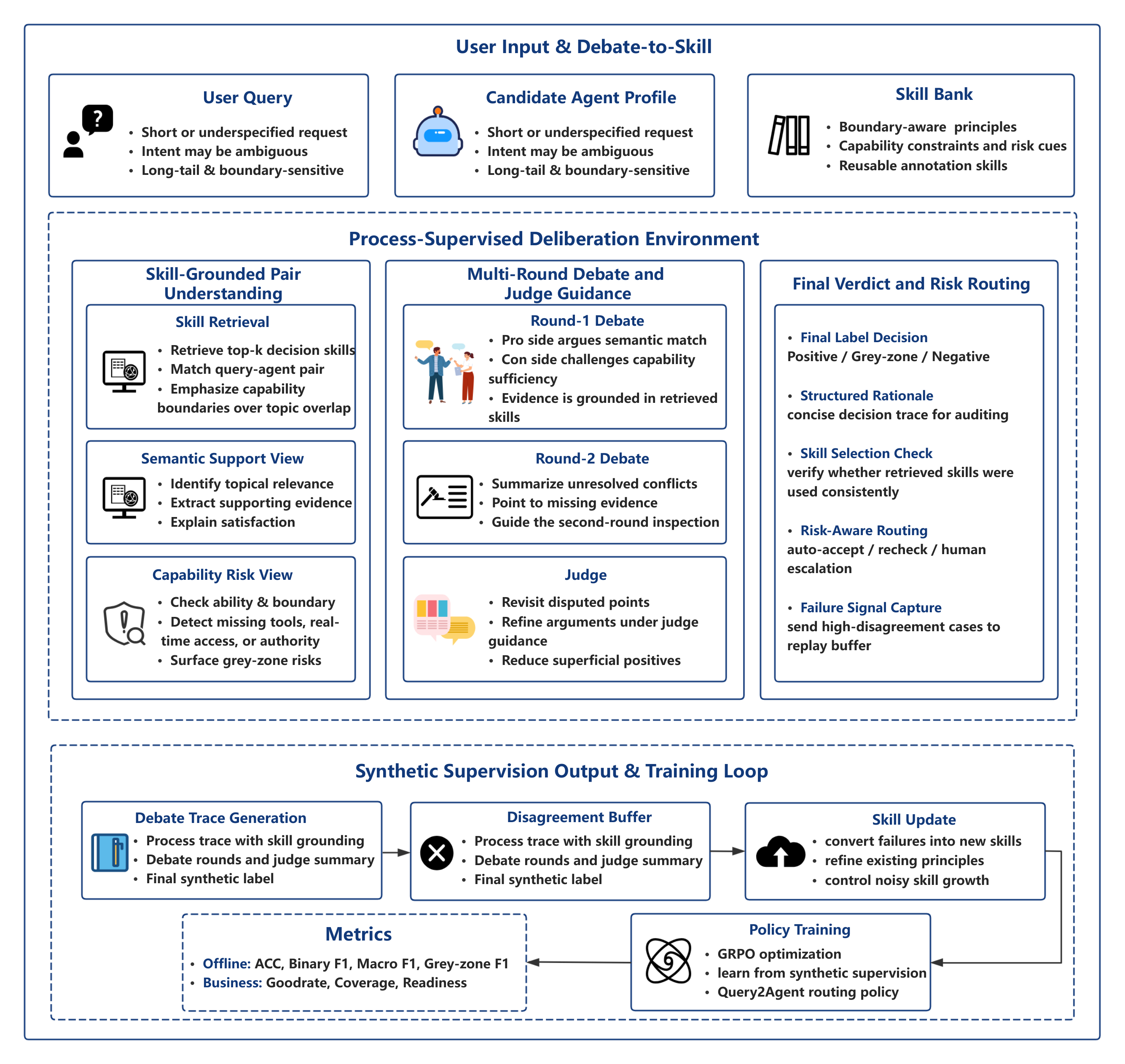}
\caption{Debate-to-Skill workflow. A query--agent pair retrieves relevant decision skills, undergoes judge-guided debate, and feeds high-priority failures into a disagreement buffer for skill refinement.}
\label{fig:d2s-workflow}
\end{figure*}

Debate-to-Skill supervises whether a query--agent pair remains valid after capability-bound inspection. As shown in Figure~\ref{fig:d2s-workflow}, it retrieves reusable decision principles, separates semantic support from capability objections, uses judge guidance to refine the trace, extracts a structured verdict, and stores recurring high-priority disagreements for principle refinement.

\subsection{Principle-Grounded Deliberation}

Each example is $x=(q,a)$, and $\mathcal{S}=\{s_1,\dots,s_{|\mathcal{S}|}\}$ is a principle-level skill bank. Each skill is
\begin{equation}
s_j = \bigl(t_j, p_j, a_j, g_j\bigr),
\label{eq:skill-object}
\end{equation}
where $t_j$ is a title, $p_j$ a capability principle, $a_j$ its application condition, and $g_j$ its tags. Given $x$, the system produces
\[
\mathcal{D}(x)=\bigl\{d_t^{\mathrm{sem}}, d_t^{\mathrm{crit}}, g_t\bigr\}_{t=1}^{T},
\]
where $d_t^{\mathrm{sem}}$ is semantic support, $d_t^{\mathrm{crit}}$ is a capability objection, and $g_t$ is judge guidance. The skill bank maps instance ambiguity into a compact principle space before deliberation.

Let $\phi(x)$ and $\phi(s)$ be capability-relevant representations of an input and a skill. Our implementation uses lexical bag-of-tokens features. Retrieval uses
\begin{equation}
\mathrm{sim}(x,s)=\frac{\phi(x)^\top \phi(s)}{\|\phi(x)\|_2 \, \|\phi(s)\|_2},
\label{eq:skill-retrieval}
\end{equation}
and the retrieved skill set is
\begin{equation}
\mathcal{S}_x=\operatorname{TopK}_{s \in \mathcal{S}} \ \mathrm{sim}(x,s).
\label{eq:topk}
\end{equation}
We use $k=3$ and fall back to seed skills when all scores are zero. Lexical retrieval is chosen for transparency, bounded latency, and auditability in the industrial annotation setting. For comparisons that retain dynamic skills, the same lexical retriever is used, so differences do not come from switching retrieval families. The \emph{w/o Dynamic Skills} ablation intentionally replaces sample-conditioned retrieval and update with fixed seed skills to measure the contribution of dynamic skill grounding; dense or learned skill retrieval is left to future work. Retrieved skills are injected into a fixed debate scaffold:
\begin{equation}
\bigl(d_t^{\mathrm{sem}}, d_t^{\mathrm{crit}}\bigr)
 = f_{\theta}\!\left(x, \mathcal{S}_x, h_{t-1}, g_{t-1}\right), \quad t=1,\dots,T,
\label{eq:debate}
\end{equation}
where $h_{t-1}$ is prior dialogue history. This factorizes semantic relevance and executable capability before labeling.

The skill bank is adaptive. Let $\mathcal{B}_t$ be the disagreement buffer of capability-sensitive conflicts. Skill updating is
\begin{equation}
\mathcal{S}_{t+1}=U\!\left(\mathcal{S}_t,\mathcal{B}_t\right),
\label{eq:skill-update}
\end{equation}
where $U$ promotes recurring high-priority conflicts into new or refined principles rather than storing raw traces.
The full procedural pseudocode is provided in Appendix~\ref{app:algorithm}.

\subsection{Process-Supervised Learning}

The structured trace is the learning target. We use reasoning-SFT on full traces and an RL-style view that verifies the final verdict. The verifier extracts
\begin{equation}
\tilde{y}=\mathrm{Extract}\!\left(g_{\theta}(x,\mathcal{S}_x,\mathcal{D}(x))\right)
\label{eq:extract}
\end{equation}
and assigns the reward
\begin{equation}
\begin{aligned}
R_{\mathrm{ver}}(x)
=\;& R_{\mathrm{cls}}(x) \\
&+ \lambda_{\mathrm{sec}}R_{\mathrm{sec}}(x) \\
&+ \lambda_{\mathrm{skill}}R_{\mathrm{skill}}(x),
\end{aligned}
\label{eq:verifier-reward}
\end{equation}
where
\begin{align}
R_{\mathrm{cls}}(x) &= \mathbf{1}\!\left[\tilde{y}=y\right], \\
R_{\mathrm{sec}}(x) &= \mathbf{1}\!\left[\mathcal{D}(x)\text{ contains all required sections}\right], \\
R_{\mathrm{skill}}(x) &= \mathbf{1}\!\left[\text{selected skill ids are valid and retrieved}\right].
\end{align}
In the current codebase, $R_{\mathrm{cls}}=1.0$ and the other terms contribute $0.1$ each. The verifier is intentionally scoped to capability consistency rather than free-form entailment: it checks verdict correctness, required sections, and valid retrieved skill ids. Logical faithfulness of arbitrary natural-language reasoning is instead encouraged through reasoning-SFT on full support-vs.-veto traces and summarized with the process-fidelity audit in Table~\ref{tab:process-audit}. GRPO with KL regularization optimizes
\begin{equation}
\begin{aligned}
\mathcal{L}_{\mathrm{RL}}(\theta)
=\;&-\mathbb{E}\!\left[
\min\!\left(
\begin{aligned}
&r_t(\theta)\hat{A}_t,\\
&\mathrm{clip}\!\left(
r_t(\theta),\,
1-\epsilon,\,
1+\epsilon
\right)\hat{A}_t
\end{aligned}
\right)
\right] \\
&+\beta\,\mathrm{KL}\!\left(
\pi_{\theta}\,\|\,\pi_{\mathrm{ref}}
\right),
\end{aligned}
\label{eq:grpo}
\end{equation}
where $r_t(\theta)$ is the policy ratio, $\hat{A}_t$ is induced by verifier reward, and $\pi_{\mathrm{ref}}$ is the reference policy. Reasoning-SFT uses the complete trace $z=(z_1,\dots,z_T)$:
\begin{equation}
\mathcal{L}_{\mathrm{SFT}}(\theta)
=
-\sum_{t=1}^{T}\log p_{\theta}(z_t\mid z_{<t},x).
\label{eq:sft}
\end{equation}
Together, Eqs.~\eqref{eq:grpo} and \eqref{eq:sft} make process supervision trainable.

\subsection{Risk-Aware Verdicts and Adaptation}

We extract the prediction from the judge section:
\begin{equation}
\hat{y} = g_{\theta}\!\left(x, \mathcal{S}_x, \mathcal{D}(x)\right),
\label{eq:judge}
\end{equation}
where $g_{\theta}$ aggregates the trace. Thus $\hat{y}$ is a risk-sensitive verdict conditioned on the input and retrieved capability principles, enabling well-structured cases to enter synthetic supervision and unstable cases to be rechecked.

The disagreement buffer links one-step deliberation to principle evolution by prioritizing cases where semantic evidence, capability evidence, and judge outcome conflict:
\begin{equation}
\begin{aligned}
p(x)=\;&\left|y^{\mathrm{sem}}-y^{\mathrm{crit}}\right| \\
&+\alpha \,\mathbf{1}\!\left[y^{\mathrm{human}}\neq y^{\mathrm{judge}}\right] \\
&+\beta \,\mathbf{1}\!\left[x \text{ triggers skill update}\right],
\end{aligned}
\label{eq:priority}
\end{equation}
with $\alpha=1.0$ and $\beta=0.5$. The $y^{\mathrm{human}}$ term is available only in offline human-audited annotation batches and is never used in evaluation or online inference prompts.

The update operator acts only on recurring, informative conflicts:
\begin{equation}
\begin{aligned}
G(c)=\mathbf{1}\!\Bigl[
\mathrm{freq}(c)\ge \tau_f \ \land \ \bar{p}(c)\ge \tau_p
\Bigr]=1,
\end{aligned}
\label{eq:gate}
\end{equation}
where $c$ is a conflict pattern. Once the gate passes, the principle space is expanded or refined:
\begin{equation}
\mathcal{S}_{t+1}=
\begin{cases}
    \mathcal{S}_t \cup \{s_{\mathrm{new}}(c)\}, &
    \text{if } \neg \mathrm{cov}(c,\mathcal{S}_t), \\
    \mathrm{Refine}\!\left(\mathcal{S}_t, c\right), & \text{otherwise.}
\end{cases}
\label{eq:add-refine}
\end{equation}
Here, $\mathrm{cov}(c,\mathcal{S}_t)$ indicates that an existing principle already covers conflict pattern $c$. Thus the system learns safer verdicts while refining the capability principles used by future verdicts. The update procedure remains conservative and rule-driven: examples with $p(x)\ge\tau$ enter a conflict buffer, and when a pattern appears at least $\tau_f=3$ times and its mean priority exceeds $\tau_p$, the corresponding principle is updated by revising its text, when-to-apply condition, or tags. For example, repeated over-acceptance of consulting-only agents for execution requests refines a boundary principle stating that topical match is insufficient when the agent lacks the required tools, authority, or transaction APIs.

\section{Experiments}

\subsection{Experiment Setup}

\paragraph{Synthetic-labeling pipeline overview.}
Our experiments follow the offline annotation loop in Figure~\ref{fig:baidu-workflow}. Starting from industrial Query2Agent traffic, we construct synthetic supervision with direct labels, reasoning traces, or Debate-to-Skill process traces, and then train downstream routing models. All models use Qwen2.5-14B-Instruct as the backbone. The teacher is the offline D2S pipeline, and the student is a Qwen2.5-14B-Instruct routing model trained with SFT and GRPO. SFT uses the binary target induced from the ternary annotation scheme, where raw-label-2 maps to positive and raw-label-1/0 map to negative; ternary labels are retained for trace construction and grey-zone analysis. GRPO samples 8 rollouts per example and uses normalized advantages $\hat{A}_i=(R_i-\bar{R})/\sigma_R$. The reward uses classification correctness with weight 1.0 and section-validity and skill-grounding rewards with $\lambda_{\mathrm{sec}}=\lambda_{\mathrm{skill}}=0.1$.

\paragraph{Data construction and splits.}
The training set contains 170K anonymized search-engine query--agent pairs collected from 2025Q4 production traffic, with raw-label ratio $\{2,1,0\}=5{:}2{:}3$. We use a separate randomly sampled 3,000-example validation set. Main Test, Domain, and Longtail each contain 1,000 human-labeled pairs, for 3,000 test examples in total; the grey-zone slice contains 312 raw-label-1 examples. Validation and test data use an 80/20 temporal composition, with 80\% from 2025Q4 and 20\% from 2026Q1. Splits are separated before training and evaluation: after normalization, the same query appears in only one of train, validation, or test. Exact query--agent duplicates are removed, and near-duplicate queries are clustered and deduplicated. Agent profiles are not globally deduplicated because one deployed agent can validly serve many queries. Candidate agents come from production serving logs: for each logged query, we take the top-$k$ online exposed candidates for annotation.

\paragraph{Annotation protocol.}
Internal domain annotators used a ternary label scheme: 2 denotes an executable match, 1 denotes a semantically related but capability-insufficient or risky pair, and 0 denotes an irrelevant pair. The two-annotator protocol applies to the human-labeled evaluation sets and manually audited subsets: each item in these sets was labeled by two annotators, disagreements were adjudicated, and raw agreement was 86.4\%. The 170k training labels are produced by the offline teacher pipeline rather than exhaustively double-annotated by humans.

\paragraph{Evaluation dataset and metrics.}
All methods are evaluated on the same fixed test protocol. Main Test is a random sample from online industrial Query2Agent traffic and serves as the primary production-style benchmark. Q2A focuses on everyday life-service scenarios, Domain covers professional traffic including business, industrial, document, emotional, and law categories, and Longtail isolates sparse and boundary-sensitive intents that are weakly represented in head traffic. We report Accuracy, Binary F1, Macro F1, and Grey-zone F1. Binary F1 collapses raw-label-1 and raw-label-0 into the negative class. Grey-zone F1 is a diagnostic one-vs-rest F1 computed from the raw ternary verdict before this collapse, treating raw-label-1 as the positive class.

\subsection{Offline Evaluation}

Our offline evaluation tests whether capability-bound supervision is a better learning target than direct-label supervision and generic reasoning. We define three structural ablations before Table~\ref{tab:main-results}. The comparison chain is intentionally explicit: SFT Label $\rightarrow$ SFT Reasoning $\rightarrow$ \emph{w/o Debate} $\rightarrow$ Debate-to-Skill. \emph{w/o Debate} removes the support-vs.-veto scaffold and makes a one-pass prediction, while keeping the same SFT+GRPO training loop. This is our closest RL-without-full-D2S control and tests whether the central support-vs.-veto process, rather than the shared GRPO loop alone, explains the gain. \emph{w/o Dynamic Skills} keeps the debate scaffold but uses the same fixed three seed skills for all examples, removing sample-conditioned skill retrieval and update. \emph{w/o Judge} removes final judge aggregation and uses the last debate round as the verdict, isolating the contribution of veto-aware aggregation. Table~\ref{tab:main-results} reports the main comparison on \textsc{Main Test} together with the grey-zone slice, so the same table serves both as the primary result table and as the core structural ablation of the process object.
\begin{table*}[t]
\centering
\small
\caption{Offline evaluation results on an industrial Query2Agent benchmark.}
\label{tab:main-results}
\begin{tabular}{lccccc}
\toprule
\multirow{2}{*}{Model} & \multicolumn{3}{c}{Main Test} & \multicolumn{2}{c}{Grey-zone} \\
\cmidrule(lr){2-4} \cmidrule(lr){5-6}
 & ACC & Binary F1 & Macro F1 & F1 & Support \\
\midrule
SFT Label & 81.3 & 77.9 & 74.6 & 45.8 & 312 \\
SFT Reasoning & 87.1 & 84.8 & 81.2 & 53.4 & 312 \\
Debate-to-Skill & 91.6 & 89.7 & 87.4 & 64.9 & 312 \\
w/o Debate & 88.6 & 86.1 & 82.9 & 56.7 & 312 \\
w/o Dynamic Skills & 89.4 & 87.3 & 84.4 & 59.8 & 312 \\
w/o Judge & 86.9 & 84.5 & 80.8 & 49.6 & 312 \\
\bottomrule
\end{tabular}
\end{table*}

Table~\ref{tab:pathology-greyzone} makes the diagnosis more explicit by isolating the two most important errors: over-accepting grey-zone pairs and missing true positives. Together, Tables~\ref{tab:main-results} and \ref{tab:pathology-greyzone} test whether the gain really comes from better capability-bound decisions rather than from generic output length or uniform conservatism.

\begin{table*}[t]
\centering
\small
\caption{Grey-zone pathology analysis on an industrial Query2Agent benchmark. Grey-zone Over-Accept Rate is the fraction of raw-label-1 pairs predicted as positive, and Positive Miss Rate is the fraction of raw-label-2 pairs predicted as non-match.}
\label{tab:pathology-greyzone}
\begin{tabular}{lcc}
\toprule
Model & Grey-zone Over-Accept Rate $\downarrow$ & Positive Miss Rate $\downarrow$ \\
\midrule
SFT Label & 33.7 & 9.8 \\
SFT Reasoning & 26.4 & 8.1 \\
Debate-to-Skill & 16.9 & 7.4 \\
w/o Debate & 23.5 & 7.9 \\
w/o Dynamic Skills & 21.2 & 7.6 \\
w/o Judge & 29.8 & 8.7 \\
\bottomrule
\end{tabular}
\end{table*}

\begin{table*}[!t]
\centering
\small
\caption{Offline evaluation across professional-domain and long-tail evaluation slices.}
\label{tab:trainset-results}
\begin{tabular}{lcccc}
\toprule
\multirow{2}{*}{Model} & \multicolumn{2}{c}{Domain} & \multicolumn{2}{c}{Longtail} \\
\cmidrule(lr){2-3} \cmidrule(lr){4-5}
 & ACC & F1 & ACC & F1 \\
\midrule
SFT Label & 73.4 & 74.9 & 61.0 & 68.3 \\
SFT Reasoning & 74.0 & 75.5 & 66.2 & 73.8 \\
Debate-to-Skill & 77.4 & 79.9 & 72.2 & 78.6 \\
\bottomrule
\end{tabular}
\end{table*}

\begin{table*}[!t]
\centering
\small
\caption{Process-fidelity audit.}
\label{tab:process-audit}
\begin{tabular}{lccc}
\toprule
Model & Structured Validity $\uparrow$ & Skill Grounding Rate $\uparrow$ & Capability Conflict Resolution $\uparrow$ \\
\midrule
Debate-to-Skill & 88.1 & 92.7 & 88.3 \\
w/o Debate & 81.6 & 91.2 & 72.4 \\
w/o Dynamic Skills & 97.5 & 62.8 & 79.1 \\
w/o Judge & 69.3 & 93.1 & 58.6 \\
\bottomrule
\end{tabular}
\end{table*}

\begin{table*}[!t]
\centering
\small
\caption{Online A/B test ($p<0.05$). Values are relative changes over the deployed SFT Label baseline.}
\label{tab:online-results}
\begin{tabular}{lccc}
\toprule
Model & $\Delta$ Card Impression Rate $\uparrow$ & $\Delta$ Card CTR $\uparrow$ & $\Delta$ Redirect CTR $\uparrow$ \\
\midrule
Baseline & 0 & 0 & 0 \\
Ours & +15.79\% & +7.63\% & +6.12\% \\
\bottomrule
\end{tabular}
\end{table*}

Table~\ref{tab:trainset-results} tests robustness across professional-domain and long-tail traffic. Table~\ref{tab:process-audit} complements the label-based metrics with process-level evidence. Structured Validity measures whether the output contains all required sections and a parseable final verdict; Skill Grounding Rate measures whether the selected skills are valid and belong to the retrieved candidate set; Capability Conflict Resolution measures, on a manually audited subset, whether the final verdict follows capability-critical evidence when semantic support and capability evidence disagree. The higher structured validity of w/o Dynamic Skills reflects more template-fixed outputs after removing skill updates, but its lower grounding and conflict resolution show weaker capability-sensitive reasoning. Its purpose is to show that the full method does not only output better labels, but also produces more valid, more grounded, and more capability-consistent decision traces.

\subsection{Online Evaluation}
We deploy Debate-to-Skill as an offline annotation layer. The treatment policy uses Debate-to-Skill synthetic supervision, while the baseline is the previous-quarter \emph{SFT Label} policy; online serving exposes only the trained routing model, so the debate cost is one-time offline generation rather than online serving latency. The online A/B test ran for four weeks with 10\% treatment traffic, with bucket assignment fixed before serving and the anonymized traffic bucket as the experimental unit. Significance was assessed using the platform's standard bucket-level A/B testing procedure, and the reported gains in card impression rate, card CTR, and downstream redirect CTR are significant at $p<0.05$. Due to internal reporting constraints, absolute traffic volumes, absolute business rates, and interval estimates are not disclosed; Table~\ref{tab:online-results} reports relative lifts over the deployed baseline. We also monitored launch guardrails, including serving stability and downstream redirect quality, and observed no launch-blocking degradation.

We report relative lift in card impression rate, card CTR, and downstream redirect CTR. Table~\ref{tab:online-results} reports an online A/B test on industrial card-serving traffic.

\section{Related Work}

\paragraph{Weak supervision and verifier-guided reasoning.}
Prior work touches our setting through weak supervision, reasoning verification, retrieval and routing, agent memory, and skill-augmented RL. Weak-supervision methods extract signals from imperfect teachers, noisy labels, and supervision diversity \cite{burns2024weakstrong,bayesianweak2025,shin2025weakstrong,greatmodels2025}. We use these works as motivation for strengthening the supervision object, but our setup keeps the same Qwen2.5-14B-Instruct backbone throughout; the novelty lies in the supervision object, not in a size-gap formulation. Reasoning work improves difficult decisions through chain-of-thought prompting, self-consistency, decomposition, and self-generated rationales \cite{wei2022cot,kojima2022zeroshot,wang2022selfconsistency,zhou2023leasttomost,zelikman2022star}. Tool use, search, self-refinement, debate, step-level verification, and consistency-style hallucination checks make these traces more operational and checkable \cite{yao2023react,schick2023toolformer,yao2023tot,madaan2023,du2024,lightman2024letsverify}. However, these lines of work still optimize different objects. Weak supervision mainly improves label quality, reasoning methods mainly improve the explicitness or checkability of a decision trace, and retrieval, routing, and skill-augmented RL mainly improve selection or execution efficiency. None of them is built around the industrial failure mode that matters here: a query can look topically relevant while still being non-executable because the candidate agent lacks the required tools, authority, real-time access, or service boundary. In Query-to-Agent matching, the core target is therefore not relevance, trace length, or skill reuse, but capability-consistent executability. Debate-to-Skill also draws on black-box output checking \cite{manakul2023selfcheckgpt}, but changes the supervised object: the trace is a capability-bound target optimized by SFT and verifier-based RL, not only an inference scaffold.

\paragraph{Retrieval, routing, and agent memory.}
Retrieval and routing work studies LLM reranking, query rewriting, active retrieval, and tool-augmented benchmarks \cite{sun2023chatgpt,ma2023query,jiang2023active,li2023api}. Structured RAG further interleaves retrieval with graph structure or speculative generation \cite{jin2026gfmrag}. Agent memory and self-adaptation work reuses prior workflows, memories, or generated experience for future tasks \cite{wang2024agentworkflow,amem2025,seal2025}. These methods mainly improve query--document relevance, tool execution, or future task reuse. Our disagreement buffer instead stores reusable decision principles for recurring query--agent capability failures.

\paragraph{Skill-augmented reinforcement learning.}
Recent skill+RL work builds skill libraries, retrieves them for new tasks, co-evolves them with the policy, or internalizes them into the agent \cite{xia2026skillrl,tu2026d2skill,lu2026skill0,zhu2026skill05,lin2026skillc}. Toolspace RL similarly learns context acquisition and tool execution under sparse rewards \cite{gupta2026atlas}. These methods optimize interactive task success. We instead optimize capability-bound annotation: skills are decision principles, and rewards measure verdict correctness, trace validity, and skill grounding. This distinction is tested by our dynamic-skill ablation and GRPO training view.

\section{Conclusion}

We introduced Debate-to-Skill, a process-supervised annotation framework for industrial query-to-agent matching. The central claim is that query-to-agent supervision should be capability-bound: the critical distinction is not topical relatedness, but whether a candidate agent remains executable after explicit capability-bound inspection. Debate-to-Skill instantiates this claim through reusable decision principles, structured deliberation traces, verifier-based optimization, and disagreement-driven refinement, showing that industrial annotation improves when the learned object matches capability-qualified executability.

\section{Limitations}

Our implementation has three main limitations. First, several modules are deliberately conservative for industrial auditability: skill retrieval is lexical rather than embedding-based, disagreement-driven updates remain rule-based, and risk-aware routing does not yet use calibrated abstention thresholds. These choices make the system easier to inspect but may underuse semantic retrieval, learned memory update, and calibrated uncertainty. Second, the verifier checks verdict correctness, section validity, and skill legality, but it does not fully verify the faithfulness of natural-language capability evidence. We currently encourage logical faithfulness through reasoning-SFT on support-vs.-veto traces and audit it with process-level metrics; learning a dedicated entailment verifier for the debate trace is future work. Third, the evaluation remains tied to an industrial Query2Agent setting. The full results require running the released training and scoring jobs end to end, and existing public benchmarks such as ToolBench and API-Bank do not contain our key raw-label-1 category: semantically plausible but capability-invalid pairs. Building a public grey-zone benchmark from open agent registries is therefore an important next step.

\bibliography{custom}

@inproceedings{du2024,
  title={Improving Factuality and Reasoning in Language Models through Multiagent Debate},
  author={Du, Yilun and Li, Shuang and Torralba, Antonio and Tenenbaum, Joshua B and Mordatch, Igor},
  booktitle={International Conference on Learning Representations},
  year={2024}
}

@inproceedings{madaan2023,
  title = {Self-Refine: Iterative Refinement with Self-Feedback},
  author = {Madaan, Aman and Tandon, Niket and Gupta, Prakhar and others},
  booktitle = {Advances in Neural Information Processing Systems},
  volume = {36},
  pages = {46534--46594},
  year = {2023}
}

@inproceedings{schick2023toolformer,
  title = {Toolformer: Language Models Can Teach Themselves to Use Tools},
  author = {Schick, Timo and Dwivedi-Yu, Jane and Dessi, Roberto and others},
  booktitle = {Advances in Neural Information Processing Systems},
  volume = {36},
  pages = {68539--68551},
  year = {2023}
}

@inproceedings{asai2024selfrag,
  title = {{Self-RAG}: Learning to Retrieve, Generate, and Critique through Self-Reflection},
  author = {Asai, Akari and Wu, Zeqiu and Wang, Yizhong and Sil, Avirup and Hajishirzi, Hannaneh},
  booktitle = {International Conference on Learning Representations},
  year = {2024}
}

@inproceedings{kojima2022zeroshot,
  title = {Large Language Models are Zero-Shot Reasoners},
  author = {Kojima, Takeshi and Gu, Shixiang Shane and Reid, Machel and Matsuo, Yutaka and Iwasawa, Yusuke},
  booktitle = {Advances in neural information processing systems},
  volume = {35},
  pages = {22199--22213},
  year = {2022}
}

@article{burns2024weakstrong,
  title = {Weak-to-Strong Generalization: Eliciting Strong Capabilities with Weak Supervision},
  author = {Burns, Collin and Izmailov, Pavel and Kirchner, Jan and others},
  journal = {arXiv preprint arXiv:2312.09390},
  year = {2023}
}

@inproceedings{wang2022selfconsistency,
  title     = {Self-Consistency Improves Chain of Thought Reasoning in Language Models},
  author    = {Wang, Xuezhi and Wei, Jason and Schuurmans, Dale and Le, Quoc and Chi, Ed and Narang, Sharan and Chowdhery, Aakanksha and Zhou, Denny},
  booktitle = {International Conference on Learning Representations},
  year      = {2023}
}

@inproceedings{wei2022cot,
  title = {Chain-of-Thought Prompting Elicits Reasoning in Large Language Models},
  author = {Wei, Jason and Wang, Xuezhi and Schuurmans, Dale and others},
  booktitle = {Advances in Neural Information Processing Systems},
  volume = {35},
  pages = {24824--24837},
  year = {2022}
}

@inproceedings{zhou2023leasttomost,
  title = {Least-to-Most Prompting Enables Complex Reasoning in Large Language Models},
  author = {Zhou, Denny and Sch{\"a}rli, Nathanael and Hou, Le and Wei, Jason and Scales, Nathan and Wang, Xuezhi and Schuurmans, Dale and Cui, Claire and Bousquet, Olivier and Le, Quoc and Chi, Ed},
  booktitle = {International Conference on Learning Representations},
  year = {2023}
}

@inproceedings{zelikman2022star,
  title = {{Star}: Bootstrapping reasoning with reasoning},
  author = {Zelikman, Eric and Wu, Yuhuai and others},
  booktitle = {Advances in Neural Information Processing Systems},
  volume = {35},
  pages = {15476--15488},
  year = {2022}
}

@inproceedings{lightman2024letsverify,
  title = {Let's Verify Step by Step},
  author = {Lightman, Hunter and Kosaraju, Vineet and Burda, Yura and Edwards, Harri and others},
  booktitle = {International Conference on Learning Representations},
  year = {2024}
}

@inproceedings{wei2022flan,
  title={Finetuned Language Models Are Zero-Shot Learners},
  author={Wei, Jason and Bosma, Maarten and Zhao, Vincent Y. and Guu, Kelvin and Yu, Adams Wei and Lester, Brian and Du, Nan and Dai, Andrew M. and Le, Quoc V.},
  booktitle={International Conference on Learning Representations},
  year={2022}
}

@inproceedings{yao2023react,
  title = {{ReAct}: Synergizing Reasoning and Acting in Language Models},
  author = {Yao, Shunyu and Zhao, Jeffrey and Yu, Dian and Du, Nan and Shafran, Izhak and Narasimhan, Karthik and Cao, Yuan},
  booktitle = {International Conference on Learning Representations},
  year = {2023}
}

@inproceedings{yao2023tot,
  title = {Tree of Thoughts: Deliberate Problem Solving with Large Language Models},
  author = {Yao, Shunyu and Yu, Dian and Zhao, Jeffrey and Shafran, Izhak and Griffiths, Thomas L. and Cao, Yuan and Narasimhan, Karthik R.},
  booktitle = {Advances in Neural Information Processing Systems},
  volume = {36},
  pages = {11809--11822},
  year = {2023}
}

@inproceedings{bayesianweak2025,
  title = {Bayesian {WeakS-to-Strong} from Text Classification to Generation},
  author = {Cui, Ziyun and Zhang, Ziyang and Sun, Guangzhi and Wu, Wen and Zhang, Chao},
  booktitle = {International Conference on Learning Representations},
  pages = {29777--29794},
  year = {2025}
}

@article{greatmodels2025,
  title = {Great Models Think Alike and This Undermines {AI} Oversight},
  author = {Goel, Shashwat and Struber, Joschka and Auzina, Ilze and others},
  journal = {arXiv preprint arXiv:2502.04313},
  year = {2025}
}

@inproceedings{jin2026gfmrag,
  title = {Generalizing Graph Foundation Models via Hyperbolic Retrieval-Augmented Generation},
  author = {Jin, Yifan and Ji, Qirui and Qin, Bin and others},
  booktitle = {Proceedings of the 32nd ACM SIGKDD Conference on Knowledge Discovery and Data Mining},
  pages = {2204--2214},
  year = {2026}
}

@inproceedings{wang2024agentworkflow,
  title = {Agent Workflow Memory},
  author = {Wang, Zora Zhiruo and Mao, Jiayuan and Fried, Daniel and Neubig, Graham},
  booktitle = {International Conference on Machine Learning},
  year = {2025}
}

@inproceedings{amem2025,
  title = {{A-Mem}: Agentic Memory for {LLM} Agents},
  author = {Xu, Wujiang and Liang, Zujie and Mei, Kai and Gao, Hang and Tan, Juntao and Zhang, Yongfeng},
  booktitle = {Advances in Neural Information Processing Systems},
  volume = {38},
  pages = {17577--17604},
  year = {2025}
}

@inproceedings{seal2025,
  title = {Self-Adapting Language Models},
  author = {Zweiger, Adam and Pari, Jyothir and Guo, Han and Kim, Yoon and Agrawal, Pulkit},
  booktitle = {Advances in Neural Information Processing Systems},
  volume = {38},
  pages = {74084--74115},
  year = {2025}
}

@inproceedings{shin2025weakstrong,
  title = {Weak-to-Strong Generalization through the Data-Centric Lens},
  author = {Shin, Changho and Cooper, John and Sala, Frederic},
  booktitle = {International Conference on Learning Representations},
  volume = {2025},
  pages = {22039--22077},
  year = {2025}
}

@inproceedings{li2023api,
  title = {{API}-Bank: A Comprehensive Benchmark for Tool-Augmented {LLM}s},
  author = {Li, Minghao and Zhao, Yingxiu and Yu, Bowen and Song, Feifan and Li, Hangyu and Yu, Haiyang and Li, Zhoujun and Huang, Fei and Li, Yongbin},
  booktitle = {Proceedings of the 2023 Conference on Empirical Methods in Natural Language Processing},
  pages = {3102--3116},
  year = {2023}
}

@inproceedings{ma2023query,
  title = {Query Rewriting in Retrieval-Augmented Large Language Models},
  author = {Ma, Xinbei and Gong, Yeyun and He, Pengcheng and Zhao, Hai and Duan, Nan},
  booktitle = {Proceedings of the 2023 Conference on Empirical Methods in Natural Language Processing},
  pages = {5303--5315},
  year = {2023}
}

@inproceedings{jiang2023active,
  title = {Active Retrieval Augmented Generation},
  author = {Jiang, Zhengbao and Xu, Frank and Gao, Luyu and Sun, Zhiqing and Liu, Qian and Dwivedi-Yu, Jane and Yang, Yiming and Callan, Jamie and Neubig, Graham},
  booktitle = {Proceedings of the 2023 Conference on Empirical Methods in Natural Language Processing},
  pages = {7969--7992},
  year = {2023}
}

@inproceedings{manakul2023selfcheckgpt,
  title = {{SelfCheckGPT}: Zero-Resource Black-Box Hallucination Detection for Generative Large Language Models},
  author = {Manakul, Potsawee and Liusie, Adian and Gales, Mark J. F.},
  booktitle = {Proceedings of the 2023 Conference on Empirical Methods in Natural Language Processing},
  pages = {9004--9017},
  year = {2023}
}

@inproceedings{sun2023chatgpt,
  title = {Is {ChatGPT} Good at Search? Investigating Large Language Models as Re-Ranking Agents},
  author = {Sun, Weiwei and Yan, Lingyong and Ma, Xinyu and Wang, Shuaiqiang and Ren, Pengjie and Chen, Zhumin and Yin, Dawei and Ren, Zhaochun},
  booktitle = {Proceedings of the 2023 Conference on Empirical Methods in Natural Language Processing},
  pages = {14918--14937},
  year = {2023}
}

@article{xia2026skillrl,
  title={SkillRL: Evolving Agents via Recursive Skill-Augmented Reinforcement Learning},
  author={Xia, Peng and Chen, Jianwen and Wang, Hanyang and Liu, Jiaqi and Zeng, Kaide and Wang, Yu and Han, Siwei and Zhou, Yiyang and Zhao, Xujiang and Chen, Haifeng and others},
  journal={arXiv preprint arXiv:2602.08234},
  year={2026}
}

@article{tu2026d2skill,
  title = {Dynamic Dual-Granularity Skill Bank for Agentic {RL}},
  author = {Tu, Songjun and Xu, Chengdong and Zhang, Qichao and Zhang, Yaocheng and Lan, Xiangyuan and Li, Linjing and Dong, Li and Zhao, Dongbin},
  journal = {arXiv preprint arXiv:2603.28716},
  year = {2026}
}

@article{lu2026skill0,
  title = {{SKILL0}: In-Context Agentic Reinforcement Learning for Skill Internalization},
  author = {Lu, Zhengxi and Yao, Zhiyuan and Wu, Jinyang and Han, Chengcheng and Gu, Qi and Cai, Xunliang and Lu, Weiming and Xiao, Jun and Zhuang, Yueting and Shen, Yongliang},
  journal = {arXiv preprint arXiv:2604.02268},
  year = {2026}
}

@article{zhu2026skill05,
  title = {{Skill0.5}: Joint Skill Internalization and Utilization for Out-of-Distribution Generalization in Agentic Reinforcement Learning},
  author = {Zhu, Jiapeng and Yu, Jianxiang and Zhao, Yibo and Han, Chengcheng and Gu, Qi and Cai, Xunliang and Li, Xiang and Qian, Weining},
  journal = {arXiv preprint arXiv:2605.28424},
  year = {2026}
}

@article{lin2026skillc,
  title = {{SkillC}: Learning Autonomous Skill Internalization in {LLM} Agents via Contrastive Credit Assignment},
  author = {Lin, Hongxiang and Kuai, Zhirui and Xue, Erpeng and Wang, Lei},
  journal = {arXiv preprint arXiv:2605.27899},
  year = {2026}
}

@inproceedings{gupta2026atlas,
  title = {Scaling Agentic Capabilities, Not Context: Efficient Reinforcement Finetuning for Large Toolspaces},
  author = {Gupta, Karan and Vajreshwari, Pranav and Pandya, Yash and Nambi, Akshay},
  booktitle = {ICLR 2026 Workshop on Reliable Autonomy},
  year = {2026}
}

\end{document}